\documentclass[journal]{IEEEtran}
\usepackage{xr}
\usepackage[utf8]{inputenc}
\usepackage{graphicx}
\usepackage{amsmath}
\usepackage{adjustbox}
\usepackage{makecell}
\usepackage{algorithm2e}
\usepackage{xcolor}
\usepackage{tikz} 
\usetikzlibrary{positioning,arrows.meta,quotes,bayesnet,matrix}
\usepackage{xcolor}
\usepackage{caption}
\usepackage{subcaption}
\usepackage{capt-of}
\usepackage{bbm}
\usepackage{hyperref}
\usepackage{amssymb}
\usepackage{stackrel}
\usepackage{pifont}
\usepackage[numbers]{natbib}

\def\tablename{Table}

\begin{document}

\title{Classifier Weighted Mixture models}

\author{Elouan Argouarc'h, Fran\c{c}ois Desbouvries \IEEEmembership{Senior Member, IEEE}, \'Eric Barat, Eiji Kawasaki, Thomas Dautremer 
\thanks{Elouan Argouarc'h and Fran\c{c}ois Desbouvries are with SAMOVAR, Télécom SudParis, Institut Polytechnique de Paris, 91120 Palaiseau, France (e-mail: elouan.argouarch,francois.desbouvries@telecom-sudparis.eu).
Elouan Argouarc'h, \'Eric Barat, Eiji Kawasaki and Thomas Dautremer are with Université Paris Saclay, CEA, List F-91120 Palaiseau, France (e-mail: elouan.argouarch, eric.barat, eiji.kawasaki, thomas.dautremer@cea.fr)}}
\markboth{ Vol. xxx, No. xxx, xxx 2021}
{Shell \MakeLowercase{\textit{et al.}}: Bare Demo of IEEEtran.cls for IEEE Journals}
\maketitle

\begin{abstract}
This paper proposes an extension of standard mixture stochastic models, by replacing the constant mixture weights with functional weights defined using a classifier. \emph{Classifier Weighted Mixtures} enable straightforward density evaluation, explicit sampling, and enhanced expressivity in variational estimation problems, without increasing the number of components nor the complexity of the mixture components.

\end{abstract}

{\color{black} 
}
\IEEEpeerreviewmaketitle
\section{Introduction}
Mixture distributions $ \sum_{k=1}^K \pi_k p_k(x)$ with $\pi_k \in (0,1)$ and $ \sum_{k=1}^K \pi_k = 1$ are powerful tools in probabilistic modeling. For instance, mixture of Gaussians distributions are widely used for density estimation and implicit clustering \cite{celeux1992classification}, because their structure makes them well suited for representing distributions with several modes, and because of the availability of a potent closed-form Expectation-Maximization (EM) algorithm \cite{dempster1977maximum}\cite{wu1983convergence}. Moreover, mixtures are a particular type of latent variable model (LVM) where the latent variable is discrete and finite. As a result, not only can mixture distributions be sampled from easily, but they also benefit from a tractable probability density function (PDF), making them versatile tools for variational estimation problems. 

Modern machine learning problems involve increasingly complex target distributions, which, from a variational estimation perspective, highlights the challenge of ensuring sufficient expressivity in the class of approximating surrogate distributions. This leads us to the limitations of classical mixtures. On the one hand, a mixture of $K$ simple distributions (e.g. Gaussians) enables to get arbitrarily close in theory \cite{stergiopoulos2000gaussian} to a given distribution, provided $K$ is large enough; but in practice, a high number of components leads to computational burdens and numerical instability. On the other hand, one can increase the expressiveness of a mixture by considering more sophisticated component distributions (denoted by $\mathcal{P}_k$). For instance, \cite{postels2021go} investigated mixtures of Normalizing-Flows (NF), where each mixture component is the result of a complex invertible mapping of some base 
distribution (denoted $\mathcal{P}_0$). Even though this approach has been successfully applied in practice on density estimation tasks \cite{yang2023mixture}, building a mixture of $K$ differents NFs is theoretically equivalent to applying a unique mapping to a base distribution which itself is a $K$-mixture, so the advantage of this approach remains to be investigated. 

Our aim in this paper is to extend the expressivity of classical mixtures, while maintaining the critical computational features of a tractable PDF and a straightforward sampling scheme. Rather than increasing $K$ or considering complex components $\mathcal{P}_k$, our approach consists in replacing the constant mixture weights $\pi_k$ by functions $\pi_k(x)$. In practice, we will achieve this goal by using a classifying function, yielding a new approach for building parameterized distributions which we refer to as \emph{Classifier Weighted Mixtures} (CWM). The paper is organized as follows. First, in section \ref{construction}, we detail the construction of the CWM PDF and unravel the underlying latent variable structure. Next, in section \ref{application}, we discuss the parameterization and applications to variational estimation problems, and compare our approach to alternative models through simulations \footnote{\textcolor{black}{Code available at \href{https://github.com/ElouanARGOUARCH/Classifier-Weighted-Mixtures}{https://github.com/ElouanARGOUARCH/Classifier-Weighted-Mixtures}}}. 

\section{From standard mixtures to CWM}\label{construction}

Our goal is to extend a mixture by replacing the constant weights $\pi_k \in (0,1)$ by functions $\pi_k: x \rightarrow \pi_k(x)$:
\begin{equation}\label{extension_objective}
    \sum_{k=1}^K \pi_k p_k(x) \longrightarrow  \sum_{k=1}^K \pi_k(x) p_k(x).
\end{equation}
\begin{figure}[h]
\includegraphics[width=\linewidth]{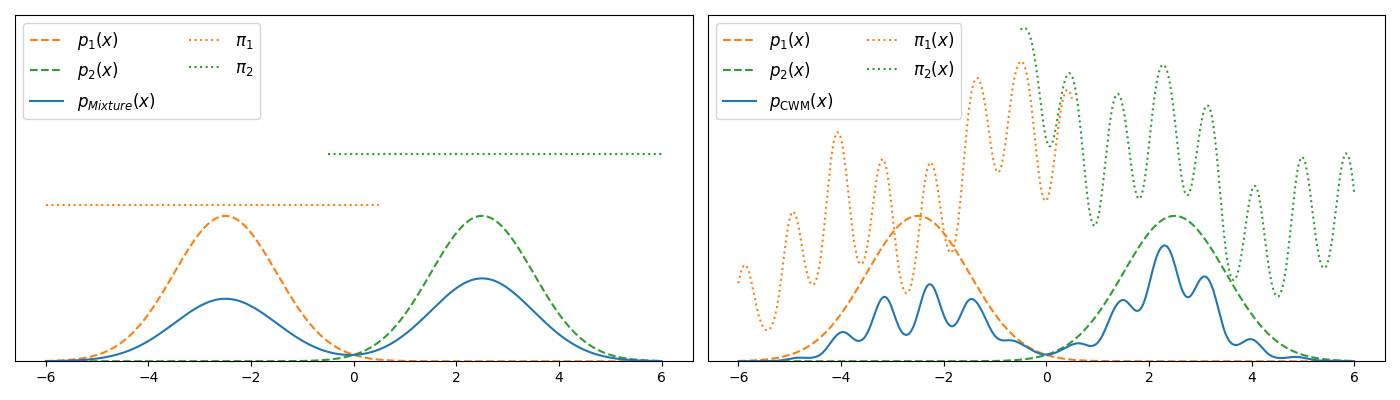}
\caption{Classical mixtures (left) can be extended to mixtures with functional and non-necessarily constant weights (right).}\label{figure_objective}
\end{figure}

Of course, these functions cannot be arbitrary. In section \ref{valid_construction}, we thus propose sufficient conditions on $\pi_k(x)$ such that $\sum_{k=1}^K \pi_k(x) p_k(x)$ is a valid PDF, which leads us to the CWM solution. In section \ref{LVM}, we unravel its underlying latent variable structure, which shows that CWM can easily be sampled from. Finally in section \ref{parameterization}, we decompose our stochastic model into its three fundamental components.

\subsection{A classifier based construction}\label{valid_construction}

\begin{figure}[h]
\includegraphics[width=\linewidth]{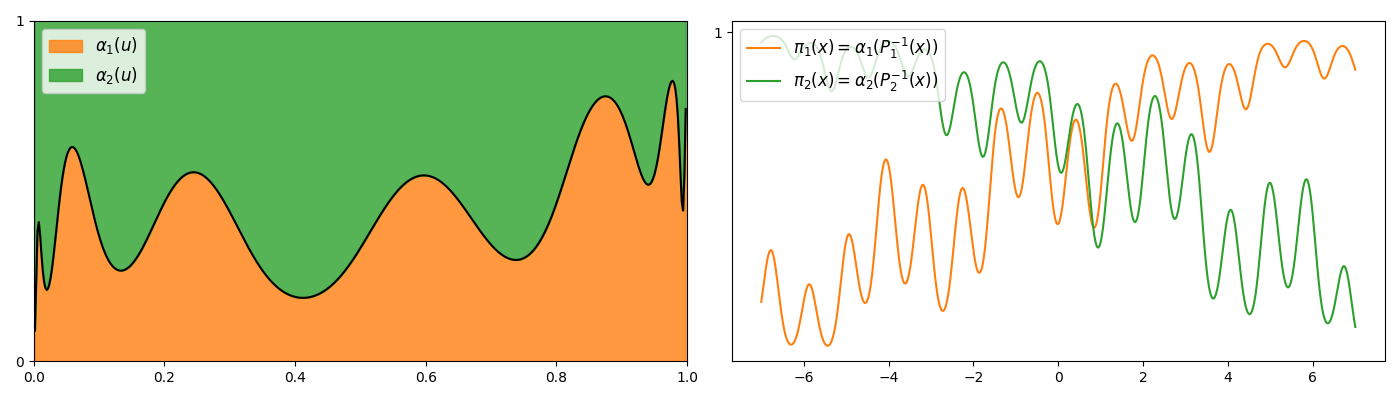}
\caption{Classifier $\alpha_k(u)$ and corresponding weights $\pi_k(x)$.}\label{figure_weights}
\end{figure}

Let us focus on the right hand side of \eqref{extension_objective}. For it to be a valid PDF, this function needs to be non-negative and sum to $1$. We therefore aim to constrain functions $\pi_k(x)$ such that:
\begin{eqnarray}
    &\sum_{k=1}^K \pi_k(x) p_k(x) \geq 0; \label{positive_condition}
    \\
    &\int\sum_{k=1}^K \pi_k(x) p_k(x)\mathrm{d}x = 1\label{integral1_condition}.
\end{eqnarray}
Firstly, \eqref{positive_condition} is satisfied if all addends $\pi_k(x) p_k(x) \geq 0$, and thus if \fbox{$\pi_k(x)\geq 0$}. To fulfill the second condition, we use $P_k$ the cumulative distribution function (CDF) of $\mathcal{P}_k$. With the change of variable $u = P_k(x)$, we rewrite the left-hand side of \eqref{integral1_condition} as:
\begin{eqnarray}
    &\int \sum_{k=1}^K \pi_k(x) p_k(x) \mathrm{d}x 
    = \int_{[0,1]} \sum_{k=1}^K \pi_k(P_k^{-1}(u))\mathrm{d}u. 
\end{eqnarray}
If \fbox{$\sum_{k=1}^K \pi_k(P_k^{-1}(u)) = 1$}, the previous integral reduces to $1$ and \eqref{integral1_condition} is satisfied. We have hence identified sufficient conditions on $\pi_k(x)$ such that $\sum_{k=1}^K \pi_k(x) p_k(x)$ is indeed a PDF. Finally, to see how a classifier is involved, we set $\alpha_k(u)\stackrel{\Delta}{= }\pi_k(P_k^{-1}(u))$, which enables us to rewrite the two previous conditions about $\pi_k(x)$ as:
\begin{eqnarray}
\label{condition-alpha}
    &\sum_{k=1}^K \alpha_k(u) = 1 \text{ and } \alpha_k(u) \geq 0 \text{ for all } u\in [0,1].
\end{eqnarray}
It is now clear that functions $\alpha_k(u)$ define a vector of probabilities, so these functions are nothing but a classifier. Therefore, with such a $K$-label classifying function, the construction:
\begin{equation}\label{result_pdf}
    \sum_{k=1}^K \underbrace{\alpha_k(P_k(x))}_{\pi_k(x)}p_k(x) \stackrel{\Delta}{=} p_{\textrm{CWM}}(x);
\end{equation} 
is indeed a valid PDF. Note that $\sum_{k=1}^K \alpha_k(u) = 1$, but $\sum_{k=1}^K \pi_k(x) \neq 1$: weights $\pi_k(x)$ are defined via a classifier (whence the name CWM), but, unless all functions are constant (in which case $p_{\textrm{CWM}}(x)$ reduces to a classical mixture), are not themselves a classifier. In figure \ref{figure_weights}, we display the classifying functions $\alpha_k(u)$ and the corresponding weights $\pi_k(x)$ which produced the CWM distribution in figure \ref{figure_objective}. 

\subsection{The underlying latent variable construction}\label{LVM}

In the previous section we built a mixture weighted by functions $\pi_k(x)$ built from classifying functions $\alpha_k(z)$. We now explain that, even though our construction is more intricate than a classical mixture model, it can nonetheless be described as a latent variable model. Therefore, in addition to a straightforward PDF evaluation mechanism (see equation \eqref{result_pdf}), the resulting distribution also benefits from an explicit sampling scheme. 

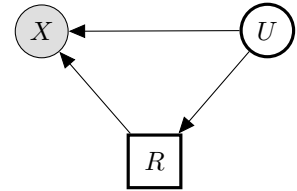
\begin{figure}[h!]
\centering
\begin{tikzpicture}[transform shape, node distance=1cm, roundnode/.style={circle, draw=black, very thick, minimum size=5mm},squarednode/.style={rectangle, draw=black, very thick, minimum size=7mm}][!ht]
        \node[squarednode](R){$R$};%
        \node[obs](X)[above= of R, xshift = -1.5cm]{$X$};%
        \node[roundnode][above=of R,xshift=1.5cm](U){$U$}; %
        \edge {R,U}{X}  
        \edge{U}{R}
    \end{tikzpicture}
\caption{Directed graph associated with the CWM construction}\label{graph_construction}
\end{figure}

PDF \eqref{result_pdf} is expressed as a sum over values $k=1,...,K$, and is thus the marginal over a couple $(X,R)$, where $R$ is categorical, and their joint PDF are the added terms in \eqref{result_pdf}: 
\begin{equation}\label{joint_x_R}
    p_\text{CWM}(x,R=k) = \alpha_k(P_k(x))p_k(x).
\end{equation}
As it stands, this joint PDF is not factorized as a marginal over $(R)$ multiplied by a conditional $(X|R)$ (unlike in usual mixture models). However, augmenting this joint PDF with the additional variable $u\in[0,1]$ and applying the change of variable $X = P_k^{-1}(U)$ yields a PDF over $(X,R,U)$ which indeed is factorized as $(U)$ times $(R|U)$ times $(X|R,U)$: 
\begin{equation}\label{full_joint}
    p_\text{CWM}(x,R=k,u) = \underbrace{\vphantom{\delta_{P_k^{-1}(u)}}\mathbbm{1}_{[0,1]}(u)}_{(U)}\underbrace{\vphantom{\delta_{P_k^{-1}(u)}}\alpha_k(u)}_{(R|U)}\underbrace{\vphantom{\delta_{P_k^{-1}(u)}}\delta_{P_k^{-1}(u)}(x)}_{(X|R,U)}.
\end{equation}
This joint PDF has an appropriate factorization and we can hence deduce the corresponding directed graph (see figure \ref{graph_construction}).

From this directed graph, it is easy to deduce a sampling scheme. Indeed, LVMs benefit from built-in procedures: by following the directed graph we can sequentially sample the latent variables, and then sample the observed variable. In our case, we can obtain samples from this distribution with: 
\begin{align}
    &U \sim \text{ Uniform}([0,1]),
    \\
    &R \sim \text{ Categorical}(\alpha(U)),
    \\
    & X = P_{R}^{-1}(U),
\end{align}
where $\alpha(U)=[\alpha_1(U),...,\alpha_K(U)]^T$.

\subsection{Change of base distribution and practical construction}
\label{parameterization}
In section \ref{valid_construction}, we proposed a construction based on a uniform base distribution, and on the CDFs $P_k$ of the mixture components $\mathcal{P}_k$. But indeed we can replace the uniform distribution by another one without additional conditions about $\pi_k(x)$. To see this, let us replace $U$ by $Z$ $\sim \mathcal{P}_0$. This corresponds to augmenting graph \ref{graph_construction} into:

\begin{equation*}
\vcenter{\hbox{\begin{tikzpicture}[transform shape, node distance=1cm, roundnode/.style={circle, draw=black, very thick, minimum size=5mm},squarednode/.style={rectangle, draw=black, very thick, minimum size=7mm}][!ht]
        \node[squarednode](R){$R$};%
        \node[obs](X)[above= of R, xshift = -.75cm]{$X$};%
        \node[roundnode][above=of R,xshift=.75cm](U){$U$}; %
        \node[roundnode][above=of R,xshift=2.25cm](Z){$Z$}; %
        \edge {R,U}{X}  
        \edge{U}{R}
        \edge{Z}{U}
    \end{tikzpicture}}}
\qquad\qquad
\begin{aligned}
    &Z \sim \mathcal{P}_0
    \\
    &U = P_0(Z)
    \\
    &R \sim \text{Categorical}(\alpha(U))
    \\
    & X = P_{R}^{-1}(U)
\end{aligned}
\end{equation*}

If we denote $w_k(z) \stackrel{\Delta}{=} \alpha_k(P_0(z))$, then $w_k(z)$ and $\alpha_k(u)$ are both equivalently classifying functions. On the other hand, the inverse CDFs $P_k^{-1}$ (resp. $P_0\circ P_k^{-1}$) are simply used as a change of variable in order to transform the base distribution $\text{Uniform}([0,1])$ (resp. $\mathcal{P}_0$) into $\mathcal{P}_k$. So in practice (see section \ref{invertible_mappings_practice} below) in order to adjust these mixture components, we shall consider invertible parameterized mappings $T_k$. With $\mathrm{J}_{T_k}$ 
denote the Jacobian matrix of $T_k$, the CWM PDF finally reads: 
\begin{equation}
\label{parameterized_pdf}
    p_\text{CWM}(x) = \sum_{k=1}^K w_k(T_k(x))p_0(T_k(x))|\text{det}(\mathrm{J}_{T_k}(x))|.
\end{equation}

\section{Application to variational estimation problems}\label{application}

We now apply our CWM model to variational problems. In Section \ref{param-application} we build our model step-by-step, considering issues such as sampling, PDF computation, and reparameterization. In the mean time, we detail the construction that we used in our experiments, in which, because we want to highlight the potential expressivity induced by functional weights $\pi_k(x)$ as compared to constant ones $\pi_k$, we purposefully use simple Gaussian mixture components (see \ref{result_param}). Next in section \ref{variational} we apply our model in the context of variational problems. Section \ref{experiments} displays simulation results.

\subsection{Parameterization}
\label{param-application}
\subsubsection{Base distribution}\label{base_distribution_practice}

Remember from section \ref{parameterization} that we can theoretically start from any parameterized base distribution.  However, if we need to sample from the corresponding CWM model  \eqref{parameterized_pdf} (resp. evaluate its PDF), the base distribution must be sampled from easily (resp. benefit from a closed-form PDF). Moreover, application to variational estimation problems might call for a reparameterization of the gradient procedure (see section \ref{reparameterization_gradients}). In turn this means that the base distribution should benefit from a sampling procedure which can be reparameterized, enabling to propagate gradients with respect to the parameters of the base distribution through its samples. 

Now, it is common practice in LVMs \cite{blei2003latent}\cite{goodfellow2014generative}\cite{kingma2014autoencoding} to consider parameter-free base distributions; moreover, in our experiments, we will focus on comparing our model to a Gaussian mixture model (GMM) and to NFs (see section \ref{experiments}). For all these reasons, we choose a standard, parameter-free, normal distribution:
\begin{equation}\label{base_distribution_param}
    \mathcal{P}_0 = \mathcal{N}(0,\mathrm{I}).
\end{equation}

\subsubsection{Classifying function}\label{classifier_practice}
Recall that it is sufficient that functions $w_k$ compute positive values which add to $1$; therefore, any classifying function works \cite{nelder1972generalized}\cite{lecun1998gradient}. In this paper we experiment with a fully-connected feed-forward Neural Network (NN), say $f$, which outputs $K$ logit values: 
\begin{equation}\label{classifier_param}
    w(z) = \text{Softmax}(f(z)).
\end{equation}
The weights and biases parameters will be adjusted during the variational estimation procedure, see section \ref{variational}.

\subsubsection{Invertible mapping}\label{invertible_mappings_practice}
We use invertible mappings $T_k$ to transform the base distribution $\mathcal{P}_0$ into the mixture components with PDF given by $p_0(T_k(x))|\det \mathrm{J}_{T_k}(x)|$, and which can be sampled from with $T_k^{-1}(z), z\sim \mathcal{P}_0$. We can parameterize these invertible mappings using NF architectures (see  \cite{rezende2015variational}\cite{kingma2016improved}\cite{dinh2017density}\cite{papamakarios2017masked}\cite{durkan2019neural} or \cite{kobyzev2020normalizing}\cite{papamakarios2021normalizing} for reviews of NF methods). In our experiments, we aim to highlight the advantage of using functional weights $\pi_k(x)$ rather than constant weights $\pi_k$, so we build a mixture of simple Gaussians with: 
\begin{equation}\label{invertible_mappings_param}
    T_k(x) = \Sigma_k^{-1/2}(x-\mu_k);
\end{equation}
where means $\mu_k$ and covariance matrices $\Sigma_k$ are parameters to be adjusted during the optimization. Moreover in practice, to ensure that $\Sigma_k$ remain positive-definite during gradient updates, we can consider diagonal matrices and parameterize the coefficients using unconstrained log-variance parameters. 

\subsubsection{A resulting model}\label{result_param}

Specific choices of base distribution, classifying function and invertible mappings, as discussed above, lead to different architectures. For instance, in our simulations (see section \ref{experiments}), we will use the construction based on \eqref{base_distribution_param}, \eqref{classifier_param} and \eqref{invertible_mappings_param}; in that case equation \eqref{parameterized_pdf} reduces to:
\begin{equation}\label{CWM_gaussians_pdf}
    p_\text{CWM}(x) = \sum_{k=1}^Kw_k(\Sigma_k^{-1/2}(x-\mu_k))\mathcal{N}(x;\mu_k,\Sigma_k). 
\end{equation}
It is indeed a mixture of Gaussian distributions, but weighted by a NN classifier function. 
\begin{equation}\label{CWM_gaussians}
\vcenter{\hbox{\begin{tikzpicture}[transform shape, node distance=1cm, roundnode/.style={circle, draw=black, very thick, minimum size=5mm},squarednode/.style={rectangle, draw=black, very thick, minimum size=7mm}][!ht]
        \node[squarednode](R){$R$};%
        \node[obs](X)[above= of R, xshift = -1cm]{$X$};%
        \node[roundnode][above=of R,xshift=1cm](Z){$Z$}; %
        \edge {R,Z}{X}  
        \edge{Z}{R}
    \end{tikzpicture}}}
\qquad\qquad
\begin{aligned}
    &Z \sim \mathcal{N}(0,\mathrm{I})
    \\
    &R \sim \text{Categorical}(w(Z))
    \\
    & X = \Sigma_R^{1/2}Z + \mu_R
\end{aligned}
\end{equation}
    
\subsection{Application to Variational estimation problems}\label{variational}

We now see how CWM can be used for solving variational estimation problems. We focus on two types of problems: density estimation in section \ref{density_estimation}, and reparameterization gradients in section \ref{reparameterization_gradients}. Let $\theta$ denote the set of all parameters (of the base distribution $\mathcal{P}_0$, of the weight function $w_k(.)$ and of the invertible mappings $T_k$), and let $p_\theta(x)$ the resulting CWM PDF (see equation \eqref{parameterized_pdf}). 

\subsubsection{Density Estimation}\label{density_estimation}

In density estimation, we dispose of recorded samples $\mathcal{D} = \{x_1,..., x_{|\mathcal{D}|}\}$ from a distribution of interest, say $\mathcal{P}$, and the goal is to build a surrogate distribution $\mathcal{P}_{\theta^*}$ that closely resembles $\mathcal{P}$ by finding appropriate parameters $\theta^*$. In practice, this is usually achieved with maximum likelihood estimation (MLE) (which is equivalent to minimizing a Monte Carlo (MC) estimate of the Kullback-Leibler divergence between $\mathcal{P}$ and $\mathcal{P}_\theta$) approximated using gradient ascent. In this case, the log-likelihood of $\mathcal{D}$, $\sum_{i=1}^{|\mathcal{D}|} \log(p_\theta(x_i))$, is differentiable with respect to $\theta$ and we can thus solve the MLE problem by gradient ascent (at least approximately towards a local maxima). 

Now, when considering the specific parameterization discussed in the previous section (see \eqref{CWM_gaussians_pdf} \eqref{CWM_gaussians}), forcing the classifier $w_k(z)$ to constant values $\pi_k \in (0,1)$ (which can be achieved by setting all the weights of the last layer of the classifier $f$ to $0$ and its biases to $\log(\pi_k)$) results in a standard mixture of Gaussians. Therefore, we can easily use the EM algorithm for Gaussian mixtures as a pre-training step, and next use the (constant weights) mixture produced by that EM algorithm as the initial point of the gradient-based optimization of the CWM model, in which the classifier $w_k(z)$ will no longer predict constant values. 

\subsubsection{Reparameterization Gradients}\label{reparameterization_gradients}

Many variational tasks \cite{sutton1999policy}\cite{kingma2014autoencoding}\cite{ganin2016domain} can be expressed as minimizing over $\theta$, by gradient descent, expectation $\mathbbm{E}_{X\sim \mathcal{P}_\theta}(h(X))$, where $h$ is a measurable function. It is therefore of interest to compute, or at least estimate with MC, the gradient of $\mathbbm{E}_{X\sim \mathcal{P}_\theta}(h(X))$ with respect to $\theta$. However, computing the gradient of the crude estimate:
\begin{equation}
    \mathbbm{E}_{X\sim \mathcal{P}_\theta}(h(X)) \approx \frac{1}{M}\sum_{i=1}^M h(x_i), x_i\stackrel{iid}{\sim}\mathcal{P}_\theta,
\end{equation}
is difficult since the samples depend on parameters $\theta$.

Practitioners often rely on reparameterization tricks \cite{kingma2014autoencoding}\cite{figurnov2018implicit} that consist of rewriting samples using a bivariate function, differentiable with respect to $\theta$, and that detangles the randomness of the parameters. We now discuss the {\sl reparameterization gradients} of a CWM model. 

First, as we have mentioned in section \ref{parameterization}, it is necessary to parameterize the base distribution appropriately in this context, and we suppose that one can reparameterize its samples in order to propagate the gradients with respect to parameters of the base distribution. But, since sampling from a CWM involves the discrete variable $R$, samples $X = T_R^{-1}(Z)$ cannot be reparameterized using a differentiable function because $R = \sum_{k=1}^K \mathbbm{1}(U\leq \sum_{j=1}^k w_j(Z))$, $U\sim \mathcal{U}_{[0,1]}$ is not continuous and we cannot propagate the gradients through this variable. Besides the reinforce approach \cite{williams1992simple} which is suited for both discrete and continuous variables (but leads to high variance estimates \cite{xu2019variance}), notorious workarounds include the Gumbell-Softmax \cite{jang2017categorical}\cite{balog2017lost} for categorical variables. It consists in approximating $R = \arg\max_{k}(g_k + \textrm{log}(w_k(Z))), g_k\stackrel{iid}{\sim}\text{Gumbell}(0,1)$ by replacing the argmax operator with a $\text{softmax}$, which yields a differentiable, but biased estimate.

In our case, since the categorical variable is a latent variable, we will show that a differentiable (approximate) reparameterization of $R$ is not required. Instead, we can build an estimate where this variable is marginalized out, following the Rao-Blackwellization (RB) \cite{casella1996rao} principle. RB is based on $\mathbbm{E}(h(X)) = \mathbbm{E}(\mathbbm{E}(h(X)|Z))$ \cite{blackwell1947conditional}. Since we can compute 
    $\mathbbm{E}(h(X)|Z) = \sum_{k=1}^K w_k(Z)h(T_k^{-1}(Z))$,
only the outer expectation calls for an MC approximation. This leads to the estimate:

\begin{equation}\label{RaoBlackwellDKL}
\mathbbm{E}_{X\sim \mathcal{P}_\theta}(h(X)) \approx \frac{1}{M}\sum_{\stackrel{i=1}{z_i\sim \mathcal{P}_0}}^{M}\underbrace{\sum_{k=1}^K w_k(z_i)h(T_k^{-1}(z_i))}_{\mathbbm{E}(h(X)|Z=z_i)}.
\end{equation}

The interest of \eqref{RaoBlackwellDKL} is twofold. First, $\mathbbm{V}\text{ar}(\mathbbm{E}(h(X)|Z))  = \mathbbm{V}\text{ar}(h(X)) - \mathbbm{E}(\mathbbm{V}\text{ar}(h(X)|Z))$ \cite{rao1992information}, so \eqref{RaoBlackwellDKL} has lower variance than the crude estimate. Next, computing $\mathbbm{E}(h(X)|Z)$ marginalizes out $R$, so \eqref{RaoBlackwellDKL} does not involve sampling from this variable; the estimate is thus differentiable and no longer relies on a reparameterization of the categorical samples.

\subsection{Experiments}\label{experiments}

\begin{figure}[!ht]
    \centering
    \begin{subfigure}[b]{0.31\linewidth}
        \includegraphics[width=\linewidth]{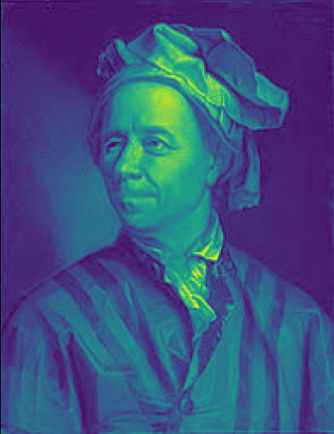}
        \caption{2D image density}
        \label{fig:image}
    \end{subfigure}
    \begin{subfigure}[b]{0.31\linewidth}
        \includegraphics[width=\linewidth]{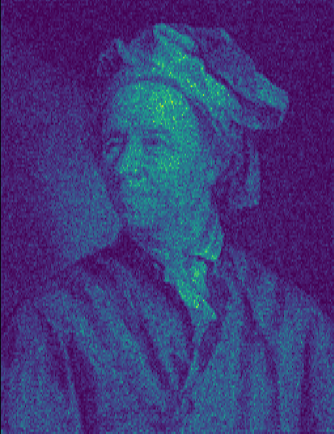}
        \caption{samples (hist.)}
        \label{fig:samples}
    \end{subfigure}
    \begin{subfigure}[b]{0.31\linewidth}
        \includegraphics[width=\linewidth]{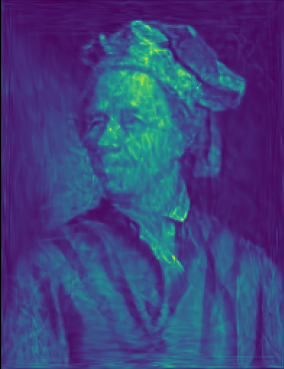}
        \caption{CWM}
        \label{fig:CWM}
    \end{subfigure}
    
    \begin{subfigure}[b]{0.31\linewidth}
        \includegraphics[width=\linewidth]{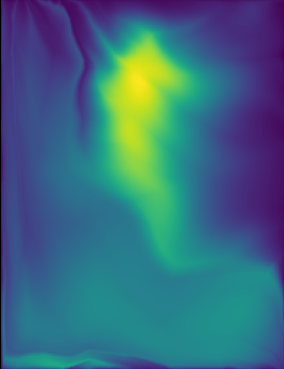}
        \caption{RNVP}
        \label{fig:RNVP}
    \end{subfigure}
    \begin{subfigure}[b]{0.31\linewidth}
        \includegraphics[width=\linewidth]{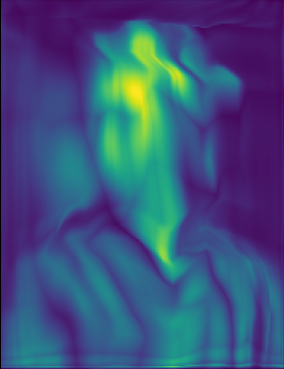}
        \caption{NSF}
        \label{fig:NSF}
    \end{subfigure}
    \begin{subfigure}[b]{0.31\linewidth}
        \includegraphics[width=\linewidth]{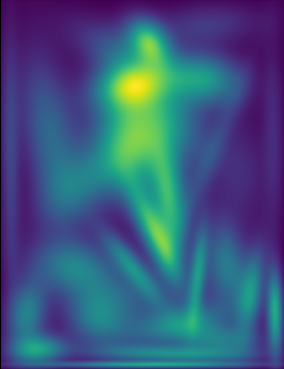}
        \caption{GMM}
        \label{fig:EM}
    \end{subfigure}
    \caption{Image - samples - CWM - RNVP - NSF - GMM \label{fig:euler_experiment}}
\end{figure}

In this illustrative section, we compare our CWM model to NF architectures (RNVP \cite{dinh2017density}, NSF \cite{durkan2019neural}) and to GMMs on a density estimation task. We draw samples (b) (an histogram is displayed in the figure) from the 2-dimensional distribution associated with a grey-scale image (a); for NFs and CWM, the density is estimated via gradient ascent of the log-likelihood (after pre-training, as mentioned in section \ref{density_estimation}); for GMM we used the EM algorithm. We summarize our results by displaying the PDF of each model (compared to the original image) in figure \ref{fig:euler_experiment}, and the likelihood scores in table \ref{log_likelihood_scores}.

\begin{table}[!ht]
\begin{adjustbox}{width=\linewidth,center}
\begin{tabular}{|l|c|c|c|c|}
   \hline
    {} & {CWM $(K=50)$} & {RNVP} &  {NSF} & {GMM $(K=50)$} \\
   \hline
    {\small \# Parameters} & {$39925$} & {$85780$} & {$79560$} & {$343$} \\
   \hline
   {\small Likelihood} & {$\mathbf{0.221\pm1.6\mathrm{e}-3}$} & {$0.133\pm 1.1\mathrm{e}-2$} & {$0.182\pm 3.9\mathrm{e}-3$} & {$0.150\pm 2.2\mathrm{e}-3$}\\
   \hline
\end{tabular}
\end{adjustbox}
\caption{Architectures, parameters and log-likelihood}\label{log_likelihood_scores}
\end{table}
Our approach outperforms the other architectures in terms of likelihood score, and with fewer number of parameters (at least compared to NF models). This is confirmed visually: the resulting PDF closely resembles the original image, which means that our stochastic model can easily  represent distributions with disjoint elements of mass, fine details and sharp edges.

\section{Conclusion}

In this paper we propose to extend classical mixtures of distributions, where the weights are constant, by mixtures where the weights become functional. More precisely, the CWM solution we propose is a construction where the functional weights are built using a classifying function. This approach allows the use of flexible (possibly NN-based) classifiers, increasing the expressiveness of our stochastic model in variational estimation problems, as compared to standard mixtures models, while still ensuring a tractable PDF and an explicit sampling scheme. As such, CWM models are suited for density estimation tasks but also allow for reparameterization gradients using RB of its discrete latent variable. Finally, our experiments showcase the interest of our approach compared to other architectures.

\bibliographystyle{ieeetr}
\bibliography{bibliography}

\end{document}